\UseRawInputEncoding
\documentclass[3p]{elsarticle}

\usepackage{lineno,hyperref}
\usepackage{amsmath}
\usepackage{amssymb}
\usepackage{longtable}
\usepackage{lipsum}  
\usepackage{color,soul}

\DeclareMathOperator*{\argmax}{argmax} 
\modulolinenumbers[5]

\journal{Elsevier Neural Networks}

\begin{document}

\begin{frontmatter}

\title{Reinforcement Learning and its Connections with Neuroscience and Psychology}

\author[cnrladdress,eeedeptaddress]{Ajay Subramanian\corref{mycorrespondingauthor}\fnref{fn1}}
\ead{f20170371@goa.bits-pilani.ac.in}

\author[cnrladdress,eeedeptaddress]{Sharad Chitlangia\fnref{fn2}}
\ead{f20170472@goa.bits-pilani.ac.in}

\author[cnrladdress,biodeptaddress]{Veeky Baths\corref{mycorrespondingauthor}}
\ead{veeky@goa.bits-pilani.ac.in}

\address[cnrladdress]{Cognitive Neuroscience Lab, BITS Pilani K K Birla Goa Campus, NH-17B, Zuarinagar, Goa, India 403726}
\address[eeedeptaddress]{Dept. of Electrical and Electronics Engineering, BITS Pilani K K Birla Goa Campus, NH-17B, Zuarinagar, Goa, India 403726}
\address[biodeptaddress]{Dept. of Biological Sciences, BITS Pilani K K Birla Goa Campus, NH-17B, Zuarinagar, Goa, India 403726}

\cortext[mycorrespondingauthor]{Corresponding author}
\fntext[fn1]{Ajay Subramanian is currently at New York University.}
\fntext[fn2]{Sharad Chitlangia is currently at Amazon.}




\begin{abstract}
Reinforcement learning methods have recently been very successful at performing complex sequential tasks like playing Atari games, Go and Poker. These algorithms have outperformed humans in several tasks by learning from scratch, using only scalar rewards obtained through interaction with their environment. While there certainly has been considerable independent innovation to produce such results, many core ideas in reinforcement learning are inspired by phenomena in animal learning, psychology and neuroscience. In this paper, we comprehensively review a large number of findings in both neuroscience and psychology that evidence reinforcement learning as a promising candidate for modeling learning and decision making in the brain. In doing so, we construct a mapping between various classes of modern RL algorithms and specific findings in both neurophysiological and behavioral literature. We then discuss the implications of this observed relationship between RL, neuroscience and psychology and its role in advancing research in both AI and brain science.
\end{abstract}


\end{frontmatter}


\section{Introduction}

Reinforcement learning (RL) methods have been very successful on a variety of complex sequential tasks such as Atari \cite{mnih_playing_2013}, Go \cite{silver_mastering_2016}, Poker \cite{heinrich_smooth_2015} and Dota-2 \cite{openai_dota_2019}, often far exceeding human-level performance. Though a large portion of these successes can be attributed to recent developments in deep reinforcement learning, many of the core ideas employed in these algorithms derive inspiration from findings in animal learning, psychology and neuroscience. There have been multiple works reviewing the correlates of reinforcement learning in neuroscience \cite{lee_neural_2012, niv2009reinforcement, botvinick2020deep}. In 2012, Lee et al. \cite{lee_neural_2012} reviewed several works reporting evidence of classical reinforcement learning ideas being implemented within the neural networks of the brain. Many commonly used building blocks of RL such as value functions, temporal difference learning and reward prediction errors (RPEs) have been validated by findings from neuroscience research, thus making reinforcement learning a promising candidate for computationally modeling human learning and decision making. \\

Since 2012 however, unprecedented advancement in RL research, accelerated by the arrival of deep learning has resulted in the emergence of several new ideas apart from the classical ideas for which neuroscience analogues had earlier been found. Relatively newer research areas like distributional RL \cite{bellemare_distributional_2017}, meta RL \cite{wang_prefrontal_2018, duan_rl2_2016}, and model-based RL \cite{sutton_integrated_1990} have emerged, which has motivated work that seeks and in some cases finds, evidence for similar phenomena in neuroscience and psychology. In this review, we have incorporated these works, thus providing a well rounded and up-to-date review of the neural and behavioral correlates for modern reinforcement learning algorithms. \\

For this review, we employ the following structure. In Section \ref{sec:classical_rl}, we provide a brief overview of classical reinforcement learning, its core, and the most popular ideas, in order to enable the uninformed reader to appreciate the findings and results discussed later on. Then, in Section \ref{sec:building_blocks} we discuss some of the building blocks of classical and modern RL: value functions, reward prediction error, eligibility traces and experience replay. While doing so, we discuss phenomena from neuroscience and psychology that are analogous to these concepts and evidence that they are implemented in the brain. Following this, in Section \ref{sec:algorithms} we discuss some modern RL algorithms and their neural and behavioral correlates: temporal difference learning, model-based RL, distributional RL, meta RL, causal RL and Hierarchical RL. Having explored all of these topics in considerable depth, we provide a mapping between specific reinforcement learning concepts and corresponding work validating their involvement in animal learning (Table \ref{tab:summary-table}). Finally in Section \ref{sec:discussion}, we present a discussion on how research at the intersection of these fields can propel each of them forward. To do so, we discuss specific challenges in RL that brain science might hold key insight to, and vice versa.\\

The following two organizational choices govern our presentation of this review.

\begin{itemize}
    \item \textbf{Two-way discussion}: We present an  exhaustive review that simultaneously discusses important recent research, on how both neuroscience and psychology have influenced RL. Additionally, unlike previous work \cite{botvinick_reinforcement_2019, lee_neural_2012, botvinick2020deep}, we also discuss how ideas from RL have recently impacted research in brain science.
    \item \textbf{Modularity}: From a computational perspective, decision making can be broken down conceptually into several capabilities. For example: planning, hierarchy, valuing choices, learning to learn etc. These happen to be distinct sub-fields of reinforcement learning (model-based RL, hierarchical RL, meta RL etc.). Therefore we divide our discussion into sections, each devoted to a specific module. This modular structure makes it easier to simultaneously describe, understand and compare ideas from RL, neuroscience and psychology.
\end{itemize}

\section{Reinforcement Learning: Background}
\label{sec:classical_rl}

\begin{figure}[ht]
    \centering
    \includegraphics[width=0.75\textwidth]{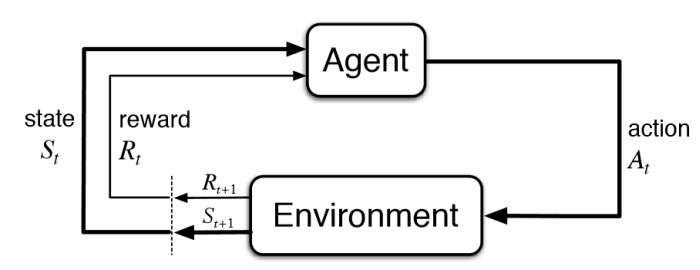}
    \caption{\textbf{The classical RL framework (Source: Sutton and Barto 1998 \cite{sutton_reinforcement_1998})}. The agent selects an action $A_t$ at state $S_t$, in response to which it receives a corresponding reward $R_{T+1}$. The objective of the agent is to choose actions that maximize its reward over a long sequence of transitions.}
    \label{fig:rl_sutton}
\end{figure}

The classical reinforcement learning framework describes an agent (human, robot etc.) interacting with its environment and learning to behave in a way that maximizes reward \cite{sutton_reinforcement_1998}. Figure \ref{fig:rl_sutton} illustrates this interaction. The agent is given a \textbf{state} $S_t$ by the environment at a time $t$. The agent, using an internal \textbf{policy} $\pi(S_t)$ or strategy selects an \textbf{action} $A_t$. The action when applied to the environment moves the agent to a new state $S_{t+1}$ and returns to it a scalar \textbf{reward} $R_{t+1}$. This sequence makes up a single \textbf{transition}. An agent's interaction with its environment comprises several such transitions. While considering these transitions, an explicit assumption is made that the future is independent of the past given the present. In other words, the next state is dependent only on the current state, action and environment properties, not on any states or actions previously taken. This is known as a Markov assumption and the process is therefore a \textbf{Markov Decision Process} (MDP).\\

In a reinforcement learning problem, the objective of the agent is to maximize the reward obtained over several transitions. In other words, the aim of the agent is to find a policy which when used to select actions, returns an optimal reward over a long duration. The most popular version of this maximization objective is to maximize discounted reward.

\begin{align}
    \pi_{\text{optimal}}(S_t) = \argmax_{\pi} \mathbb{E} \left[\sum_{\tau = t}^{\infty} \gamma^{\tau - t} R(S_{\tau}, \pi(S_{\tau})\right]
\end{align}

where $\pi_{optimal}$ is the optimal policy and $\gamma \mid 0 < \gamma < 1$ is a factor to discount future rewards. It should be noted that maximizing the reward for each transition independently might not yield an optimal long-term reward. This aspect introduces several complexities in arriving at an optimal solution for\ the RL problem, and motivates features such as exploration and planning in RL algorithms. \\

Many reinforcement learning algorithms use a value function as a way to assign utility to states and actions. The value function of a state $S_t$ is the expected reward that the agent is expected to receive if it starts at that state and executes a particular policy forever after. 

\begin{align}
    \label{valuefn}
    V(S_t) = \mathbb{E} \left[\sum_{\tau=t}^{\infty} \gamma^{\tau - t} R(S_{\tau}, \pi(S_{\tau})) \right]
\end{align}

where $V$ is the state value function. A value function could also be assigned to a state-action pair in which case it represents the expected reward if a specific policy is executed \textit{after} a given action is taken at the state.

\begin{align}
    Q_\pi(S_t, A_t) = \mathbb{E} \left[ R_{t+1} + \sum_{\tau=t+1}^{\infty} \gamma^{\tau - t} R(S_{\tau}, \pi(S_{\tau})) \right]
\end{align}

where $Q$ represents the value function for a state-action pair.

Value-function based RL algorithms often optimize value function estimates rather than directly optimizing policy. Once the optimal value function is learned, an optimal policy would then entail picking the highest value actions at each state. This procedure is called value iteration \cite{pashenkova_value_1996} and finds application in various modern reinforcement learning algorithms. A common set of algorithms for optimizing the value function are the \textbf{dynamic programming (DP) methods}. These methods update value functions by bootstrapping value functions from other states \cite{bellman_theory_1954, busoniu_reinforcement_2010}. Examples of DP methods are Q-learning \cite{watkins_q-learning_1992} and SARSA \cite{sutton_reinforcement_1998}. The optimization process involves updating the value function by ascending the gradient in the direction of the difference between \textbf{target} values and the currently estimated values, thus moving towards better estimates of rewards obtained during environment interaction. The target value is computed using DP bootstrapping. The difference between target and current value is termed as Reward Prediction Error (RPE). Dynamic programming methods that use value functions of states adjacent to the current state, to compute the target, are called temporal difference methods \cite{sutton_learning_1988} and are discussed in Section \ref{sec:td}. \\

Now that we have given a brief background on some of the important core reinforcement learning concepts: MDP, policy, value functions and dynamic programming; we will explore the neural and behavioral correlates for some of the fundamental building blocks that make up classical and modern reinforcement learning algorithms.

\section{Building Blocks of Reinforcement Learning: Neural and Behavioral Correlates}
\label{sec:building_blocks}
As briefly outlined in the previous section, the RL solution to an MDP can make use of several components or building-blocks. Among these, some of the most popular components are the value function, eligibility traces and reward prediction errors. More recently with the advent of deep RL, new components such as experience replay and episodic memory have emerged that are commonly incorporated within RL algorithms. In this section, we explore their neural and behavioral correlates. Many of these ideas have already been reviewed in much detail by Lee et al. \cite{lee_neural_2012} and so our description of them will be concise relative to topics covered in future sections of the paper. For a more in-depth review of these, see \cite{lee_neural_2012}.

\subsection{Value Functions}
As discussed in the previous section, a value function is a measure of reward expectation. This expectation is measured practically as the mean of discounted rewards over future states. 

\begin{align}
    \label{sample-valuefn}
    V(S_t) = \frac{1}{N} \sum_{\tau=t}^{t+N-1} \gamma^{\tau - t} R(S_{\tau}, \pi(S_{\tau}))
\end{align}
where N is the number of sample states considered.\\

Neural signals containing information about reward expectancy have been found in many areas of the brain \cite{schultz_neuronal_2000, hikosaka_basal_2006, wallis_heterogeneous_2010}. Resembling the two types of value functions prevalent in RL algorithms, evidence has been found for the brain too encoding both state values and action value. Action value functions \cite{samejima2005representation} are useful during motor responses when an action needs to be selected while state value functions might play an evaluative role. Transformations between the two types have been found to occur in the brain. For instance, during decision making, state value functions transform from a mean over all actions into value functions for the chosen action, which are often referred to as \textbf{chosen values}. \cite{lau2008value, padoa-schioppa_neurons_2006,cai_heterogeneous_2011}.\\

Despite these similarities between neural value signals and value functions employed in reinforcement learning, they are different in some important ways. In RL, value functions for different decisions are all treated the same and represent the expected value of a single reward. But in the brain, activity for action value functions are observed in various areas for a single decision \cite{seo_lateral_2009, kim_prefrontal_2008, pastor-bernier_neural_2011, cai_heterogeneous_2011, so_supplementary_2010} and might apply to distinct reward signals as demonstrated by Murdoch et al. \cite{murdoch2018place} in the case of place preference (navigation) and song syllables (singing), in songbirds. They observed that strobe light flashes aversively conditioned place preference but not vocal learning, while noise bursts aversively conditioned vocal learning but not place preference. Another case of value signals in different brain regions encoding expectancy for different rewards for the same decision was seen in animals during a juice flavor decision making task \cite{padoa-schioppa_neurons_2006}. Neurons encoding value in the supplemental motor area signaled desirable eye movements (to spatial location of targets) when considering the choice while those in the primate orbitofrontal cortex were associated with the juice flavors themselves (targets) \cite{tremblay1999relative, wallis2003neuronal}. The function of distinct reward signals in the brain is analogous to a highly distributed version of backpropagation in artificial neural networks wherein distinct reward signals could selectively update distinct sets of parameters in the network.\\

Neural signals for chosen values are also distributed in multiple brain areas \cite{padoa-schioppa_neurons_2006, lau2008value, kim_role_2009, sul_distinct_2010, cai_heterogeneous_2011}. Also, some brain regions encode the difference of values between two alternative actions to determine likelihood of taking an action over another \cite{seo_behavioral_2009, kim_prefrontal_2008, pastor-bernier_neural_2011, seo_lateral_2009, cai_heterogeneous_2011}. This is similar to the use of baseline values in \textbf{actor-critic} methods where value functions of each action are scaled by their mean to obtain values relative to other actions at the state \cite{sutton1999policy}. These baseline values are used to compute the `advantage' of a given action relative to others available in the action space.

\subsection{Reward Prediction Error}
In order to make good decisions about states that are desirable to visit, we need a good estimate of the value of a state. Most reinforcement learning algorithms optimize the value function by minimizing a \textbf{reward prediction error} (RPE). If $V_\pi(S_t)$ is the value function (expected reward) at a state $S_t$ and $G_t = R_{t+1} + R_{t+2} + R_{t+3} + \ldots$ is the sum of rewards obtained after time $t$ (also known as \textbf{return}), then a common formulation of the RPE at $t$ is the difference between the two terms.

\begin{align}
    RPE(t) = G_t - V_\pi(S_t)
\end{align}

A policy that minimizes the reward prediction error for all states would return the best value estimates for all states in the state space. Thus, the RPE helps improve value estimates. Such RPE signals have been identified in midbrain dopaminergic neurons \cite{schultz_behavioral_2006} and many other areas such as the orbitofrontal cortex, lateral habenula and angular cingulate cortex \cite{matsumoto_lateral_2007, hong_globus_2008, matsumoto_medial_2007, seo_temporal_2007, sul_distinct_2010, kim_role_2009, oyama_reward_2010}.\\

Action value functions in the brain are believed to be updated and stored at the synapses between cortical axons and striatal spiny dendrites \cite{reynolds_cellular_2001, hikosaka_basal_2006, lo_cortico-basal_2006, hong_dopamine-mediated_2011}. RPEs are input to these synapses through terminals of dopaminergic neurons \cite{levey_localization_1993, schultz_behavioral_2006, haber_striatonigrostriatal_2000, haber_reward_2010} and value functions are updated.

\subsection{Credit Assignment and Eligibility Traces}
In many cases and quite commonly in human behavior, rewards for a task are temporally delayed. In other words, decisions have to be made at several states before a reward feedback is obtained. An example of this is cooking, where a complete recipe has to be executed before actually tasting the dish and determining whether it is good or bad (reward). Now, say the reward was negative, we need to be able to effectively determine the step of the recipe where we went wrong so that it can be corrected on the next trial. This is known as the \textbf{credit assignment problem}. It is the challenge of finding the ``responsibility" of each encountered state for an obtained reward. Behavioral experiments on reversal learning tasks have shown that credit assignment problems surface in animals with lesions in the orbitofrontal cortex which therefore suggests its involvement in assigning credit to states \cite{iversen_perseverative_1970, schoenbaum_orbitofrontal_2002, fellows_ventromedial_2003, murray_what_2007}.\\

In RL literature, there are two prominent techniques that have been developed to solve the credit assignment problem. The first approach is to introduce intermediate states \cite{montague_framework_1996} to strengthen the connection between a state and its corresponding reward. This technique, however, doesn't correspond to any observation in neuroscience literature and is not consistent with profiles of dopamine neuron activity \cite{pan_dopamine_2005}. The second method is to use eligibility traces which are short term memory signals that assign state responsibility for a reward \cite{sutton_reinforcement_1998}. Eligibility traces are higher for states that are on average closer to the reward. Unlike intermediate states, eligibility traces have been observed in several animals and brain regions including the prefrontal cortex, striatum and frontal cortex \cite{barraclough_prefrontal_2004, seo_behavioral_2009, seo_lateral_2009, kim_encoding_2007, kim_role_2009, sul_distinct_2010, sul_role_2011, curtis_beyond_2010}. The orbitofrontal cortex of the brain is believed to play an important role in credit assignment. Its involvement is evidenced by the observation that neurons in the orbitofrontal cortex show increased activity when a positive reward is obtained from a specific action. \cite{barraclough_prefrontal_2004, seo_behavioral_2009, kim_role_2009, roesch_ventral_2009, sul_distinct_2010, abe_distributed_2011}. Additionally, neurons in orbitofrontal cortex are believed to also encode relationships between actions and their corresponding outcomes \cite{barraclough_prefrontal_2004, seo_lateral_2009, kim_role_2009, sul_distinct_2010}. This observation could inspire future work in reinforcement learning research towards a solution to the credit assignment problem. Ongoing work is also exploring and testing the involvement of eligibility traces for credit assignment in synaptic value updates. A popular idea in this direction is neuromodulated STDP which attempts to model this by combining classical STDP with eligibility traces to add external reinforcement from reward signals \cite{shen_dichotomous_2008, gerfen_modulation_2011}.

\begin{figure}[ht]
    \centering
    \includegraphics[width=0.75\textwidth]{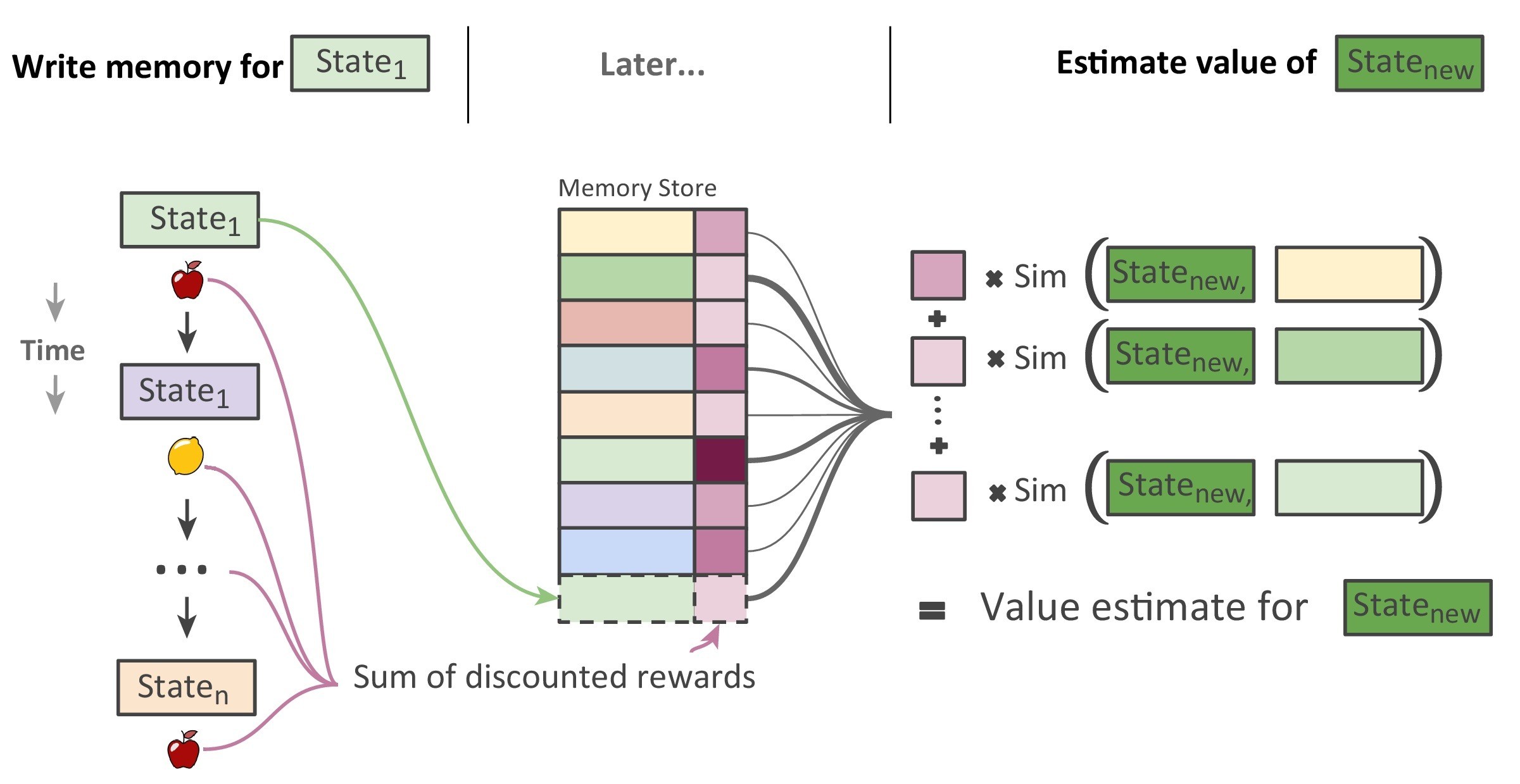}
    \caption{\textbf{Episodic reinforcement learning (Source: Botvinick et al. 2019 \cite{botvinick_reinforcement_2019})} Initially, the agent plays out a sequence of actions and stores the encountered states along with their learned value functions (expected sum of discounted rewards). When a new state is encountered in a new trajectory, its value can be estimated as linear combination of stored value functions weighted by the similarity of the stored states to the new state. This helps avoid learning value functions from scratch for every new state.}
    \label{fig:episodic}
\end{figure}

\subsection{Experience Replay and Episodic Memory}
\label{sec:experience_replay}

Rodent research has led to the discovery that place cells and grid cells in the hippocampus encode a spatial map of the environment \cite{moser_place_2008, o1971hippocampus, sargolini2006conjunctive}. Along with mapping trajectories, these cells spontaneously recap previously experienced trajectories \cite{diba_forward_2007, foster_reverse_2006, louie_temporally_2001, skaggs_replay_1996}. They can also explore new spatial trajectories which haven't been experienced before \cite{gupta_hippocampal_2010, olafsdottir_hippocampal_nodate}, a phenomenon which is known as \textbf{replay}. Replay's involvement in playing out trajectories that never happened suggest that it might be important in the brain's learning of world models \cite{foster_replay_2017, pezzulo_internally_2014} which is used to generalize learned knowledge. Biological replay mechanisms have been recorded in the entorhinal cortices \cite{olafsdottir_coordinated_2016}, prefrontal cortex \cite{peyrache2009replay} and visual cortices \cite{ji_coordinated_2007}. However, the information stored varies between these areas. Entorhinal cortical replay encodes spatial relationship between objects while visual cortical replay encodes sensory properties of events and objects. Replay mechanisms have been incorporated in modern deep reinforcement learning methods in the form of experience replay \cite{mnih_playing_2013, schaul_prioritized_2016}.\\

As discussed previously, deep RL methods that excel in performance in various tasks struggle with achieving sample efficiency similar to that of humans \cite{lake_ullman_tenenbaum_gershman_2017, tsividis_human_nodate}. Experience replay is a popular component in reinforcement learning algorithms which enables them to learn tasks with fewer environment interactions by storing and reusing previously experienced transitions to update the policy. However, experience replay mechanisms in deep RL are still unable to mimic their biological counterparts. For instance, Liu et al. \cite{liu_human_2019} show that while experience replay in RL records experience in the same sequence as they occurred, hippocampal replay does not tend to follow this `movie' sequence and rather employs an `imagination' sequence in which experienced events are replayed in the order in which they are expected to occur according to learned internal models. Thus, integration of experience replay with model-based RL is an exciting avenue for future deep RL research.\\

Additionally, experience replay in deep RL involves using only previously played trajectories of the same task that the agent is learning and hence do not assist in learning new tasks. Another approach called episodic RL \cite{pritzel_neural_2017, gershman_reinforcement_2015, lengyel_hippocampal_nodate} uses such experience as an inductive bias to learn future tasks. An illustration of an episodic RL algorithm is shown in Figure \ref{fig:episodic}. This approaches employs a similarity network which reuses values for states that have already been learned, thus reducing the time to learn values for new states \cite{lin_episodic_2018}. This idea is similar to instance-based models of memory in where specific stored information from past experience is used for decision making in new situations. \cite{gershman_reinforcement_2017, lengyel_hippocampal_nodate, bornstein_reinstated_2017, bornstein_reminders_2017}.

\section{Algorithms for Reinforcement Learning: Neural and Behavioral Correlates}
\label{sec:algorithms}
Having covered the building blocks that most commonly make up RL algorithms, we now dive deep into various types of reinforcement learning algorithms along with work in both neuroscience and psychology suggesting that they might be promising models for certain aspects of animal learning and decision making.\\

\subsection{Temporal Difference Learning}
\label{sec:td}

Temporal difference learning is one of the central ideas in reinforcement learning. The most common formulation of the reward prediction error discussed in the previous section, is the temporal difference (TD) error. The TD error $\delta_t$ is defined as:

\begin{align}
    \label{tdeq}
    \delta_t = R_{t+1} + \gamma V(s_{t+1}) - V(S_t)
\end{align}

The origins of TD learning can be traced back to the Rescorla-Wagner (RW) model for classical conditioning \cite{rescorla1972theory}, that learning occurs only when the animal is surprised. It proposed a trial-level associative learning rule which updates ``associative strengths" of stimuli using a prediction error.

\begin{align}
    \label{rweq}
    \delta_{RW} = \alpha(R_n - V(S_n))
\end{align}

where $\alpha$ is the learning rate, $R_n$ is the unconditional stimulus (US, reward) and $V(S_n)$ is the associative strength for conditional stimulus $S$ which measures how well it predicts the US. While this model explains aspects of classical conditioning such as blocking \cite{kamin1967predictability}, it is trial-based and so does not explain temporal dependencies of learning and consequently, higher-order conditioning. The TD model fills these gaps.\\

The earliest known use of temporal difference learning in artificial intelligence research dates back to 1959 when Samuel \cite{samuel_studies_1959} demonstrated its usage for a checkers-playing program. Sutton \cite{sutton_learning_1988} developed the first theoretical formulation of TD learning and showed that it was more efficient and more accurate than conventional supervised learning methods. \\

Following this, work in computational neuroscience suggested that the firing of dopamine neurons signalled a reward prediction error \cite{montague_using_1993}. Later work \cite{sejnowski_predictive_1995} showed that the TD model allows a formulation of expectations through value functions to influence synaptic changes, via a Hebbian learning framework. Later, from the extensive experiments conducted by \cite{schultz_predictive_1998}, a major breakthrough in relating TD methods to actual biological phenomena was made by Montague and colleagues \cite{montague_framework_1996} when they related fluctuation levels in dopamine delivery, from the VTA/SNc to cortical and subcortical target neuronal structures, to TD reward prediction errors. \\

The TD error formulation in RL is a very specific case of the more general TD($\lambda$) proposed by Sutton \cite{sutton_reinforcement_1998} which accounts for eligibility traces. The concept of eligibility traces was inspired from ideas like trace conditioning \cite{pavlov1927conditioned} and Hull's learning theory \cite{hull1943principles}. Two new terms are introduced to account for it, the weight vector $w_t$ and the eligibility trace $z_t$, modeled as:

\begin{align}
\delta_t &= R_{t+1} + \gamma V(S_t, w_t)  - V(S_{t}, w_{t}) \\
w_t &= w_{t-1} + \alpha \delta_{t-1} z_{t-1} \\
z_t &= \gamma \lambda z_{t-1} + \gamma V(S_{t-1}, w_{t-1})
\end{align}

When $\lambda = 1$, this formulation perfectly mimics the behavior of \textbf{Monte Carlo algorithms} and the credit given to previous steps decreases by a factor of $\gamma$. On the other hand, when $\lambda = 0$, it transforms into the TD formulation discussed earlier, where only the previous state is given credit. \cite{sutton_reinforcement_1998} showed that this TD($\lambda$) formulation is the same formulation as TD model of classical conditioning as used in the framework proposed by Montague and others \cite{montague_framework_1996} to verify the results of TD learning. Thus, the TD($\lambda$) formulation combined RPEs and eligibility traces into a single framework. \\

The TD model accounts for many limitations of the RW model. Bootstrapping of value functions (using a value estimate in the target expression) allows it to explain higher-order conditioning. It also provides flexibility in terms of real-time stimulus representation; with existing work having shown that one in particular, the microstimulus representation corresponds well with several empirical phenomena \cite{ludvig2012evaluating}. Additionally, given its temporal nature, it has been able to make predictions about aspects of animal learning some of which were confirmed later on. For instance, some of its prominent predictions \cite{sutton1981toward, sutton1990time} were: a) conditioning requires a positive inter-stimulus interval (ISI), b) remote associations facilitate conditioning, c) blocking reverses if the new CS temporally precedes pretrained CS (later verified experimentally in rabbits \cite{kehoe1987temporal}). The TD model has also been successful in modelling observations from neuroscience. Perhaps the most commonly related neuroscience phenomenon to the TD reward prediction error was given by Schultz \cite{schultz_predictive_1998} as the reward prediction error hypothesis of dopamine neuron activity. \\

The TD model is, however, limited in a few ways. Firstly, it says nothing about how responses are learned, given the learned ability to predict stimuli. Secondly, the model in its simplest form cannot express uncertainty because it uses the point-average as an estimate for future reward. Recent approaches which model the value function as a distribution, account for uncertainty \cite{dabney_distributional_2020, gershman2015unifying}. These are discussed further in Section \ref{sec:distributionalrl}.

\begin{figure}[ht]
    \centering
    \includegraphics[width=0.65\textwidth]{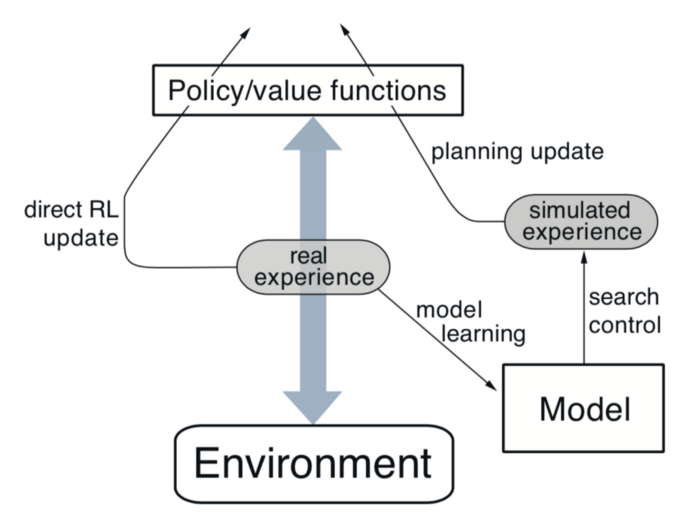}
    \caption{\textbf{Schematic of the Dyna model-based reinforcement learning framework \cite{sutton_integrated_1990}}. Involves simultaneous model-free and model-based procedures. The model-free procedure involves using interactions with the real environment ('real experience') to directly learn the policy and/or value functions. The model-based procedure uses the real experience to learn a dynamics model $P(S_{t+1},R_{t+1} \vert S_t, A_t)$ which can be used to generate artificial/simulated experience as another way to update policy and value functions.}
    \label{fig:model-based}
\end{figure}

\subsection{Model-based Reinforcement Learning}

The classical reinforcement learning framework accounts only for learning that occurs through interaction. However, a large portion of learning in humans and animals involves imagined scenarios, planning out consequences of actions, replay (as discussed in Section \ref{sec:experience_replay}) and so on. Model-based reinforcement learning seeks to mimic these capabilities and is a promising area both in RL \cite{silver_mastering_2016} (Figure \ref{fig:model-based}) and as a computational model for biological learning \cite{doll2012ubiquity}.\\

The animal learning community in the early 20th century saw a divide between Thorndike's Law of Effect \cite{Thorndike173} and Tolman's Cognitive Maps \cite{PMID:18870876}. Thorndike posited that humans associate rewards to actions and our future choices are driven by the type of reward we receive. On the other hand, Tolman stated that learning can still happen in the case that a reward is not immediately received, strengthening the argument for a type of \textbf{latent learning}, requiring goal-directed planning and reasoning. Thorndike's Law of Effect and Tolman's Cognitive Maps have served as foundational behavioral evidence for the two major types of learning systems concerned with action valuation in our brain, model-free and model-based learning.\\

Model-based learning systems involve building mental models through experience. There is evidence that model-based algorithms are implemented in biological systems. For instance, Glascher et al. \cite{glascher_states_2010} observed increased activity in the lateral prefrontal cortex when previously unknown state transitions were observed. This evidence showed that the brain integrates unknown transitions into its transition model. Additionally, the hippocampus might play a role in integrating information about the current task and behavioral context. This integration might rely on synchronous activity in the theta band of frequencies \cite{sirota_entrainment_2008, benchenane_coherent_2010, hyman_what_2011, womelsdorf_selective_2010}. Langdon et al. \cite{Langdon2018ModelbasedPF} reviewed recent findings of the association of dopaminergic prediction errors with model based learning and hypothesized that the underlying system might be multiplexing model-free scalar RPEs with model-based multi-dimensional RPEs. \\

Although there have been numerous advancements in finding neural correlates for model-free reinforcement learning \cite{doi:10.1152/jn.2000.84.6.3072, Hare2008DissociatingTR, Knutson2006LinkingNA}, the last two decades have witnessed research that bolsters evidence for the existence of a model-based system especially in a combined setting with the model-free learning system \cite{Kool2018CompetitionAC, seo_lateral_2009, Daw2011ModelbasedIO, Glscher2010StatesVR, Ito2011MultipleRA}. Human neural systems are known to use information from both model-free and model-based sources \cite{daw_uncertainty-based_2005, pan_reward_2008, glascher_states_2010}. Typical experiments involve a multi-staged decision making task while simultaneously recording BOLD (blood-oxygen-level-dependent) signals through fMRI. Results from these experiments suggest strongly coupled decision making systems (existence of reward prediction probability signals of both types of information \cite{lau2008value}) in the ventromedial prefrontal cortex \cite{tsutsui_dynamic_2016, Daw2011ModelbasedIO}, ventral striatum \cite{Daw2011ModelbasedIO} and a model based behavior in the lateral pre-frontal cortex \cite{Glscher2010StatesVR, Dolan2013GoalsAH, daw_uncertainty-based_2005} and anterior cingulate cortex \cite{Akam2021WhatID}. Combined social information and reward history can also be traced to the different regions of the anterior cingulate cortex \cite{Apps2016TheAC}.\\



Recently, the idea of \textbf{successor representations}\cite{Dayan1993ImprovingGF} has been revived, especially in the context of how humans build cognitive maps of the environment. The underlying idea behind successor representations is to build a `predictive map' of states of the environment in terms of future state \textit{occupancies}. Recent modeling work \cite{Russek2017PredictiveRC} has re-confirmed previous work on how successor representations might lie between model based and model free learning when compared over the spectrum of flexibility and efficiency.
Gershman \cite{Gershman2018TheSR} explains and summarizes recent literature on successor representations and its neural correlates. Akam and Walton \cite{Akam2021WhatID} recently proposed that successor representations are a key module in how model-based reinforcement learning systems are coupled with model-free reinforcement learning systems. Specifically, they proposed that future behaviors and RPEs are refined by imagined (or offline) planning and successor type representations to be incorporated as and when necessary (during online interaction). \\

Humans are known to develop \textit{habits}: fast decisions or action sequences over time \cite{Dolan2013GoalsAH}. Habits have been associated with model-free reinforcement learning, in particular, as responses to stimuli i.e., forming an association between an action and antecedent stimuli \cite{Balleine2010HumanAR, Miller2019HabitsWV, Dezfouli2012HabitsAS}. Additionally it was also recently observed that in the context of habit development for a set of individuals there was no involvement of model-based learning \cite{Gillan2015ModelbasedLP}. \\


The idea of building models is much more central and has a lot of implications to other aspects of human intelligence including but not limited to structure learning and social intelligence. Lake et al. \cite{lake_ullman_tenenbaum_gershman_2017} suggest that human learning systems arbitrate to tradeoff between flexibility and speed. \\

\begin{figure}[ht]
    \centering
    \includegraphics[width=\textwidth]{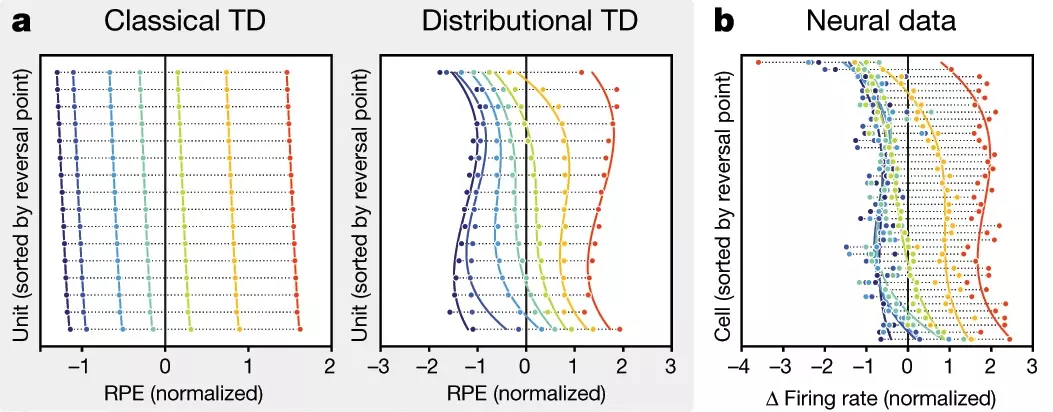}
    \caption{\textbf{Comparison of distributional TD and classical TD RPEs (Source: Dabney et al. 2020 \cite{dabney_distributional_2020})}. On each trial, the animal receives one of seven possible reward values, chosen at random. \textbf{a.} RPEs produced by classical and distributional TD simulations. Each horizontal line is one simulated neuron. Each colour corresponds to a particular reward magnitude. The x axis is the cell’s response when reward is received. In classical TD, all cells expected approximately the same RPE for a given reward signal. In contrast, in distributional TD, the simulated neurons showed significantly different degrees of reward expectation. \textbf{b.} Responses recorded from light-identified dopamine neurons in mice on the same task. A large variation in RPEs was observed between cells, which resembles distributional TD more than the classical TD simulation.}
    \label{fig:distributional}
\end{figure}

\subsection{Distributional Reinforcement Learning}
\label{sec:distributionalrl}

In classical temporal difference learning, as discussed earlier, the value of a state is the expectation of cumulative future reward. As shown by Equation \ref{sample-valuefn}, this expectation is expressed as the \textbf{mean} over future rewards. This measure is limited since it does not account for variance. Also, because the mean is a point estimator, it cannot capture multimodal return distributions. Much recent work has been done on developing a distributional framework for RL that maintains a return \textbf{distribution} rather than a single average value over future rewards \cite{bellemare_distributional_2017}. The formulation of this idea, termed as \textbf{distributional reinforcement learning}, replaces the value function $V(s_t)$ in the temporal difference update (Equation \ref{tdeq}) with $Z$, the probability distribution of returns for state $s_t$.

\begin{align}
    \delta_t = R_{t+1} + \gamma Z(s_{t+1}) - Z(s_t)
\end{align}

$Z(s_t)$ stores the probability of occurrence for each value of return possible at $s_t$. Since it stores the complete distribution of return, the expected return (value function) is simply its expectation.

\begin{align}
     V(s_t) = \mathbb{E} \left[ Z(s_t) \right]
\end{align}

Additionally, learning the return distribution helps the agent account for variance and capture multimodality. Its benefit can be immediately seen when we consider the simple example of a one step task. Suppose there are three states A, B and C. The agent starts at A and can choose to either go to B or C. Unknown to the agent: reward obtained at B is either $+1$ or $-1$ with equal probability, while reward at C is always $0.5$. Value functions learned using classical TD would learn that B and C are equivalent since expected reward is equal. On the other hand, value functions learned using distributional TD would know that B is in fact much more ``risky'' than C.\\

Practically, it is difficult to assign probabilities to every possible return value in cases where the distribution is continuous. To solve this issue, algorithms usually discretize (bin) the return distribution with the number of bins tuned as a hyperparameter. For example, the C51 model uses 51 bins to represent the return distribution for Atari games \cite{bellemare_distributional_2017}.\\

Past work had provided evidence for distributional coding in the brain for non-RL domains \cite{pouget_probabilistic_2013}. Moreover, distributional reinforcement learning had earlier been shown to be biologically plausible \cite{lammel_reward_2014, dabney_distributional_2018}. Recently, Dabney et al. \cite{dabney_distributional_2020} carried out single-unit recordings of the ventral tegmental area (VTA) in mice and showed that for a given dopamine-based reward, different cells show different Reward Prediction Errors (RPEs). These RPEs can be either positive or negative which indicates that some cells are optimistic i.e., expect a larger reward than what is obtained, while others are pessimistic and expect a lower reward. Each cell RPE here is analogous to a bin used in practical implementations of distributional RL, since the cell's spiking activity represents expectation of a specific reward prediction error. Equation \ref{neurondist_td} describes how the cell RPEs together form the $Z$ distribution seen in the RL formulation.

\begin{align*}
    \delta_{t1} &= R_{t+1} + \gamma V_1(s_{t+1}) - V_1(s_t)\\
    \delta_{t2} &= R_{t+1} + \gamma V_2(s_{t+1}) - V_2(s_t)\\
    \vdots\\
    \delta_{tN} &= R_{t+1} + \gamma V_N(s_{t+1}) - V_N(s_t)
\end{align*}
\begin{align}
    \label{neurondist_td}
    Z(s_t) &= \{ \delta_{t1}, \delta_{t2}, \dots \delta_{tN} \}
\end{align}
where $n = 1 \dots N$ represent N neurons and $\delta_n$ is the RPE of the $n$th neuron. All of these RPEs taken together form the $Z$ distribution we saw during the discussion on distributional RL. \\

Through extensive experiments, they compared the distributional coding with other models that attempt to explain RL in neural circuits, and showed that distributional RL most accurately predicts RPE reversal points and future rewards in the brain. Figure \ref{fig:distributional} shows plots comparing distributional TD and classical TD on points obtained via single cell recording.\\

Unlike many of the RL algorithms we have seen so far, distributional RL is one of the algorithms whose involvement in neural circuits was identified after the idea was first independently proposed in AI literature. Hence, it offers evidence that better and more efficient computational models can potentially result in advances in brain research.

\begin{figure}[ht]
    \centering
    \includegraphics[width=0.65\textwidth]{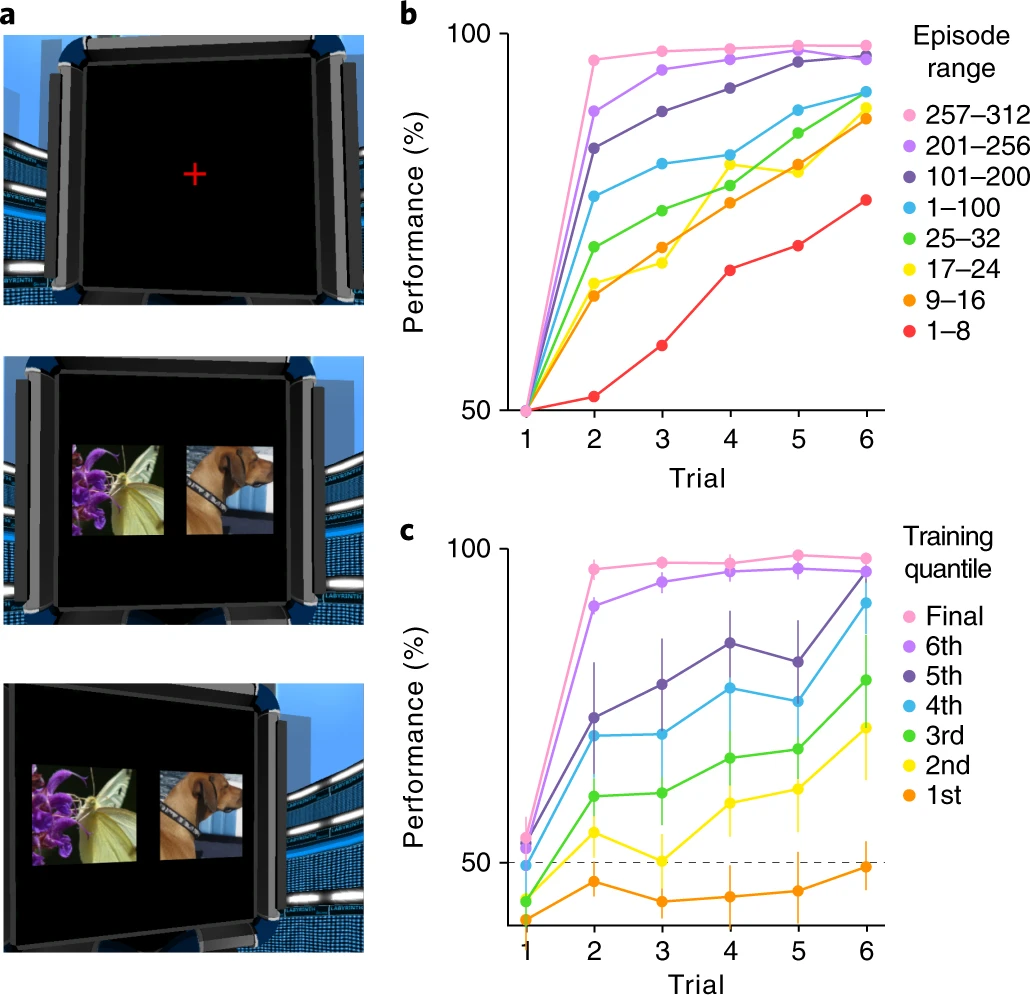}
    \caption{\textbf{Comparison between meta RL's (recurrent network) and Harlow's experimental results (Source: Wang et al. 2018 \cite{wang_prefrontal_2018})}. \textbf{a.} Inputs for simulation experiments showing fixation cross (top), initial stimulus (middle), and outcome of saccade (bottom). \textbf{b.} Performance reward gathered during each trial of Harlow's monkey experiment \cite{harlow_formation_1949}. \textbf{c.} Simulation performance at different phases of training. Performance improvement trend over time resembles Harlow's results.}
    \label{fig:meta-rl}
\end{figure}

\subsection{Meta Reinforcement Learning}
While modern deep reinforcement learning methods have been able to achieve superhuman performance on a variety of tasks, they are many orders less sample efficient than the average human and possess weak inductive biases that deter transfer of learned knowledge \cite{lake_ullman_tenenbaum_gershman_2017, marcus_deep_2018}. \\

One way to increase sample efficiency is to avoid learning tasks from scratch each time and instead use previous learning experiences to guide the current learning process. In machine learning literature, This leveraging of past experience to speed up learning of new tasks is called \textbf{meta-learning} \cite{schaul_metalearning_2010}. The original idea of ``learning to learn'' is often attributed to Harlow's 1949 work \cite{harlow_formation_1949} wherein a monkey was presented with two unseen objects, only one of which contained a reward. The monkey was then made to pick one of the objects after which the reward was revealed and the positions of the objects possibly reversed. All of this constituted a single trial. A given set of objects were used for a set of 6 trials before switching them for different objects, observing 6 trials and so on. The reward when tracked across several such rounds yielded an interesting observation. As the monkey was exposed to more sets of objects, the number of steps it needed to solve the problem for a new object set decreased. Thus, the monkey demonstrated capabilities of transferring knowledge between similar tasks. A recent idea that has helped in modelling such behavior is the hypothesis that underlying the fast learning problem of each object set, there was a slow learning process that figured out the problem dynamics and helped the monkey improve its sample efficiency on related problems \cite{botvinick_reinforcement_2019}. \\

Very recently, this core idea has been applied to reinforcement learning as meta RL to accelerate the learning process. Meta RL can be formulated as a modified version of the classical RL problem where the policy $\pi(s_t)$ now additionally depends on the previous reward $r_t$ and action $a_{t-1}$, thus becoming $\pi(s_t, a_{t-1}, r_t$). These additional dependencies allow the policy to internalize dynamics of MDP. Recurrent networks when used as part of a reinforcement learning algorithm for tasks similar to Harlow's yielded similar reward curves. This suggests that over a long period of exposure to related tasks, RNNs are able to capture the underlying activity dynamics due to their ability to memorize sequential information \cite{wang_learning_2016, duan_rl2_2016} (Figure \ref{fig:meta-rl}). Wang and colleagues \cite{wang_prefrontal_2018} also noticed that such meta-learning methods formed a part of dopaminergic reward-based learning in the prefrontal cortex of the brain. Much recent work on the prefrontal cortex (PFC) suggests that humans do more than just learning representations of expected reward. The PFC encodes latent representations on recent rewards and choices, and some sectors also encode the expected values of actions, objects, and states \cite{tsutsui_dynamic_2016, rushworth_choice_2008, padoa-schioppa_neurons_2006, kim_neural_1999, barraclough_prefrontal_2004}.\\

As an extension of meta RL, a set of recent computational approaches combines episodic memory with meta RL which results in stronger inductive biases and higher sample efficiency. Inspired from the observation that episodic memory circuits reinstate patterns of activity in the cerebral cortex \cite{ritter_been_2018}, Ritter et al. \cite{ritter_episodic_2018} developed a framework for how such episodic memory functions can strategically reuse information about previously learned tasks and thereby improve learning efficiency \cite{santoro_meta-learning_2016, wayne_unsupervised_2018}. This work also evidences recent interactions between meta model-based control and episodic memory in human learning \cite{vikbladh_episodic_2017}.

\subsection{Causal Reinforcement Learning}

The ability of humans to reason about cause and effect relationships is not unknown. The field of causal inference and reasoning looks at studying this paradigm in great detail. Pearl in 2019 \cite{10.5555/1642718} introduced a 3-level causal hierarchy which he called the ``causal ladder''. The causal ladder consists of associations, interventions and counterfactuals at the different levels of the hierarchy. Or more simply: seeing, doing and imagining respectively. \\

Much of supervised learning is at the first level i.e. learning associations from observed data. Reinforcement learning, in contrast, simultaneously has an agent learning associations and performing interventions in a world. 
Most model-free RL algorithms learn observation action associations by performing interventions in the environment. It is important to note here that to learn from interventions, an agent does not necessarily have to maintain a model of the world as long as it has access to the world or a good model of it. Humans are known to perform interventions to infer basic cause-effect relationships of the world from a very young age. It has been shown that children can learn to identify causal relationships in basic 3-variable models by performing interventions \cite{McCormack2016ChildrensUO, Sobel2010TheIO}. What this means is that, in their environments, children explore and perform interventions freely to explore and understand the world. \\

The third level of the hierarchy, which consists of counterfactuals (or imagining), usually requires an agent maintaining an internal model of the world, using which it tries to imagine counterfactual `what if?' scenarios. Being able to construct intuitive world models is a hallmark of human intelligence because it helps us in planning and reasoning among other aspects of our daily interactions. In particular, models encoding causal structure serve as strong priors for planning and reasoning about tasks \cite{lake_ullman_tenenbaum_gershman_2017}. These causal models are especially helpful in contemplating scenarios for better planning i.e., reasoning through counterfactual actions and not necessarily taking the immediately rewarding action. A counterfactual scenario can be thought of as a combination of observation and imagined interventions. In formal terms of causal inference, with a structural causal model (SCM) of the environment, counterfactual scenarios can be simulated. Recent works \cite{oberst_counterfactual_2019, buesing_woulda_2018} focus on this by modeling the environment as a structural causal model. By building an SCM, it can be intervened to take an action that was not originally taken and simulate counterfactual experience. Compared to Vanilla model-based policy search, a counterfactual policy search has been shown to perform better \cite{buesing_woulda_2018}. \\

In an experiment involving Parkinson's disease diagnosed patients, sub-second temporal resolution dopamine levels were monitored through blood-oxygen-level-dependent (BOLD) imaging to study the action of specific neurotransmitters \cite{Kishida2015SubsecondDF}. It was observed that the subsecond dopamine fluctuations encoded reward and \textit{counterfactual prediction errors} in superposition, paving the way for neural evidence of counterfactual outcomes. Much work on factual learning has suggested that humans have a valence-induced bias towards positive prediction errors over negative prediction errors \cite{Lefebvre2017BehaviouralAN, Ouden2013DissociableEO, Frank2007GeneticTD}. This means humans prefer outcomes that lead to higher reward than expected than outcomes that lead to lower rewards than expected. But as previously mentioned, an important sect of human intelligence involves learning from foregone outcomes.  Recently, it was shown that humans have a bias towards negative prediction errors rather than positive prediction errors during counterfactual learning \cite{Palminteri2017ConfirmationBI}. More generally, this means that humans have a ``confirmation bias'' towards their own choices that guides learning than either positive or negative prediction errors. Recent evidence \cite{Pischedda2020TheEO} showed that counterfactual outcomes are encoded negatively in the ventral medial prefrontal cortex and positively in the dorsal medial prefrontal cortex. Consistent with the previously described findings, factual learning is encoded in the opposite pattern in both these regions \cite{Li2011SignalsIH, Klein2017LearningRV}. More generally, experimental findings suggested that human learning behavior is significantly increased when complete information is presented (i.e., counterfactual outcomes are presented). It has also been shown that neurons in the lateral frontal polar cortex (lFPC), dorsomedial frontal cortex (DMFC), and posteromedial cortex (PMC) encode reward-based evidence favoring the best counterfactual option at future decisions \cite{Boorman2011CounterfactualCA}. \\

Building internal models of the world allows humans to deal with partial observability on a daily basis \cite{gershman_reinforcement_2017}. These internal models are grounded in concepts observed through partitioning observations in a well-organized manner i.e. structure learning from observations, which allows them to make decisions under incomplete information. In the context of reinforcement learning, algorithms depend on a representation of the environment and hence directly affect the algorithm's learning quality (efficiency and efficacy) \cite{Gershman2015DiscoveringLC}. An important component of the structure learnt is causality. In other words, the learnt structure should be able to capture the appropriate discrete causal structure underlying the continuous world \cite{Gershman2010LearningLS}. The ability of humans to build a structure consisting of \textit{latent causes} (i.e. hidden causes) from observations is remarkable yet not entirely understood and allows them to perform inference over these latent causes to reason\cite{gershman_exploring_2012}. Recent work in computational modelling supports this by building a Bayesian non-parametric prior over latent causes \cite{soto_explaining_2014}. The idea behind structure learning is central to human intelligence, even beyond the context of reinforcement learning \cite{Braun2010StructureLI}. \\
 
A standard modelling framework for building causal knowledge of the world (artificial or real) requires first identifying potential qualitative relationships (i.e., learning the structure of variables) followed by estimating their strength (i.e., quantitative cause-effect estimation). Analogously, humans tend to extract relevant information to build high-level causal knowledge and then improve the knowledge by estimating each cause's quantitative effect. As previously noted \cite{Lagnado2010BeyondCC}, much past research on causal learning and reasoning has focused on quantitative estimation of cause-effect relationships from pre-fixed variables (usually available through a dataset). However, building qualitative causal relationships is also important and foundational to recent work on ``intuitive theories'' \cite{Gerstenberg2016IntuitiveT}. As an example of building qualitative causal relationships, there are cases where we may know that two variables have a particular relationship but we may not know the extent to which they are related and which variable is \textit{causing} the other. Building causal knowledge of relevant high level variables bottom-up (e.g. from rich inputs such as pixels) is a fairly challenging task. CausalWorld \cite{Ahmed2021CausalWorldAR}, a recently proposed environment suite, involves learning the causal structure of high level variables and should enable research in this direction in the AI community. Artificial environments also already allow the ability to intervene, which is foundational to \textit{evaluating} which causal structure might be a better model of the world \cite{Hagmayer2007CausalRT, Lagnado2004TheAO}. \\

There have been various theories behind the psychology of causal induction, with the two most prominent being the causal power theory \cite{Cheng1997FromCT} and the $\Delta P$ model \cite{Lober2000IsCI}. These models address the question of how humans learn the association between causes and effects. Although these models have been quite widely argued and debated over, they fail to account for formal definitions in terms of graphical models, an idea central to modern ideas in causality \cite{10.5555/1642718}. Recently, Tenenbaum and Griffiths \cite{Tenenbaum2000StructureLI} postulated that performing inferences over learned causal structures is a central human tendency and in an attempt to bridge the psychology, computer science and philosophy literatures proposed Bayesian Causal Support and the $\chi^2$ model. Both these models extend the original causal power theory model and the $\Delta P$ model by incorporating Bayesian Inference which works on graphical models rather than simple parameter estimation.

\subsection{Hierarchical Reinforcement Learning}

General RL algorithms scale poorly with the size of state space due to difficulty in exploration and effects of catastrophic forgetting that arise in larger task domains. In order to solve this problem also known as the \textbf{scaling problem}, a popular computational framework known as temporal abstraction \cite{barto_recent_2003, dietterich_hierarchical_2000, parr_reinforcement_1998, sutton_between_1999} was developed, which suggested learning temporally extended actions that were composed of primitive actions. A common way in which these temporally extended actions are implemented is through the \textbf{options} \cite{sutton_between_1999} framework. Options are actions that span more than a single state and consist of multiple primitive actions. For example, the option "walk towards the door" would be composed of several primitive actions including motor movements and maintaining balance. Mathematically, the simplest form of options are formulated as a triplet $(\pi, \beta, \mathbb{I})$ where $\pi$ is the policy, $\beta(t) \in [0, 1]$ is the termination condition i.e. the probability that the option terminates at the current state, and $\mathbb{I}$ is the initiation set of states, which determines if the option can be chosen at the current state.\\

Hierarchical reinforcement learning combines temporally extended actions to maximize reward on goal-directed tasks. In psychology, hierarchy has played a significant role in explaining goal-directed behavior \cite{lashley_problem_1951, miller_plans_2017, newell_report_1959, anderson_integrated_2004, botvinick_doing_2004, schneider_hierarchical_2006, zacks_event_2007}. Even in neuroscience, existing literature accounts for the prefrontal cortex being largely responsible for hierarchical behavior \cite{badre_cognitive_2008, botvinick_hierarchical_2008, courtney_hierarchical_2012, fuster_theory_1989, koechlin_architecture_2003}. Thus, even though HRL was not developed to answer questions in psychology and neuroscience, it addresses an issue with standard RL methods which might also be prevalent in the brain.\\

Early work in psychology had also postulated the presence of hierarchy in human behavior. That determining the sequence of primitive actions requires higher-level representations of task context, was first formalized by Lashley in 1951 \cite{lashley_problem_1951}. The concept of task representation \cite{cohen_control_1990, cooper_contention_2000, monsell_task_2003} is very similar to the option construct (discussed earlier) that was developed in reinforcement learning literature. Empirical evidence that human mental representations are organized hierarchically was also found \cite{newtson_foundations_1976, zacks_event_2001}. Hierarchy has also been observed in the behavior of children through their childhood. Children learn elementary skills which are gradually integrated into more complex skills and knowledge as they grow \cite{bruner_organization_1973, fischer_theory_1980, greenfield_development_1972}.\\

The strongest resemblance to HRL is found in the production-system based theories of cognition, especially ACT-R \cite{anderson_integrated_2004} and Soar \cite{lehman_gentle_1996}. These frameworks propose that the solution to a problem can make use of shorter action sequences called ``chunks". Given a problem, high-level decisions can be used to trigger these chunks. Though these frameworks are similar to HRL in many regards, they differ in the aspect of not being based around a single reward maximization objective.\\

Thus, HRL shares attributes with multiple theories in behavioral psychology. However, ideas in psychology go even beyond the positive transfer problem that we have until now discussed i.e., sequencing temporally abstracted actions to develop goal-directed policies; and discuss downsides of hierarchical learning in humans. Luchins in 1942 \cite{luchins_mechanization_1942} introduced the negative transfer problem; that pre-existing knowledge with context differing from the current problem can hinder problem-solving in human subjects. Surprisingly, HRL aligns with behavioral theories even in these downsides. A direct analog to the negative transfer problem has been observed in HRL \cite{botvinick_hierarchically_2009}. \\

As a natural extension to the above-discussed work that provides strong evidence for the hierarchical nature of human behavior, recent neuroscience research has delved deeper into neural correlates for hierarchy. Ribas-Fernandes et al. \cite{ribas2011neural} showed for the first time that the medial PFC processes subgoal-related RPEs. This was followed by further study on the nature of these RPEs, which showed that subgoal-related RPEs are unsigned \cite{ribas2019subgoal}. These results reinforce recent evidence that RPEs for different task levels are induced as separable signals in the basal ganglia \cite{diuk2013hierarchical}. Balaguer et al. \cite{balaguer2016neural} extend these studies to hierarchical planning, by reporting that neural signals in the dmPFC encode the cost of representing hierarchical plans.\\

A major challenge in hierarchical RL has been option discovery, that is, how to form chunks of `reusable' actions from primitive ones. One approach to option discovery is to keep a record of states that occur frequently on paths to goals and label them as subgoals or bottleneck states that a good solution must pass through \cite{mcgovern_autonomous_2002, pickett_policyblocks_2002, thrun_finding_1994}. This bottleneck theory is also consistent with work that shows that humans are sensitive to repeating sequences of events. Another approach to option discovery from HRL literature is to construct a graph of states and all the transitions possible. Then, graph partitioning can be used to identify bottleneck states which can be used as subgoals during the learning process \cite{mannor_dynamic_2004, menache_q-cut_2002, simsek_identifying_2005}. Existing work in psychology provides empirical evidence that children identify causal representations that they then integrate into a large causal model \cite{gopnik_mechanisms_2004, gopnik_theory_2004, sommerville_infants_2005, sommerville_pulling_2005}. More recent work in HRL uses task agnostic approaches to discover options, by using intrinsic rewards for exploration \cite{barto_intrinsically_2004, singh_intrinsically_2005}. Existing neuroscience literature also provides evidence for something similar to this notion of intrinsic reward driven learning. It has been found that the same dopaminergic neurons that code reward prediction errors also respond to novel stimuli \cite{bunzeck_absolute_2006, redgrave_short-latency_2006, schultz_responses_1993}. In psychology, intrinsic rewards are strongly tied to ideas of motivation which like their RL counterpart, depend on the animal's action as well as current state. For example, a rabbit in a state of hunger is more motivated to obtain a food reward. Despite the similarity, RL algorithms still do not capture the correlation between strength/vigor of actions and motivation (which is seen in animals) \cite{skinner1988operant}, since simple environments commonly used in RL do not allow the agent to control the rate at which actions are performed. \\

A parallel line of research tries to approach the problem of hierarchy discovery, by building computational models of how the brain might do so. Solway et al. \cite{solway2014optimal} developed a Bayesian model selection approach to identify optimal hierarchies. Structure discovered using this approach explains various behavioral effects like bottleneck states, transitions and hierarchical planning. Tomov et al. \cite{tomov2020discovery} extended this approach by proposing a Bayesian model that additionally captures uncertainty-based learning and reward generalization, both of which were unexplained by previous models \cite{solway2014optimal, schapiro2013neural}.\\

\begin{figure}[ht]
    \centering
    \includegraphics[width=0.55\textwidth]{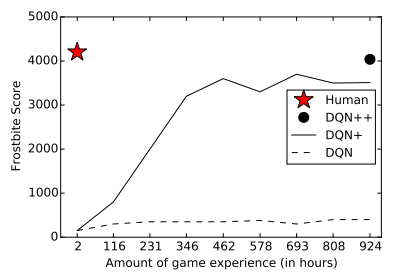}
    \caption{\textbf{Comparison of humans with state of the art deep reinforcement learning methods on the Atari game 'Frostbite' (Source: Lake et al. 2017 \cite{lake_ullman_tenenbaum_gershman_2017})}. The best Deep RL agent, DQN++ takes significantly more time ($\sim$924 hours) than humans to achieve similar performance. Other agents (DQN and DQN+) fail to match performance even after a large amount of experience.}
    \label{fig:humansvsrl}
\end{figure}

 \begin{longtable}[p]{| p{0.4\textwidth} | p{0.6\textwidth} |}
\caption{Table summarizing the mapping discussed between concepts in reinforcement learning (left column) and evidential phenomena reported in neuroscience and psychology research or specific areas of the brain responsible from them. Literature corresponding to them have also been referenced.}
\label{tab:summary-table}\\
 \hline
\textbf{Reinforcement learning concept} &
  \textbf{Corresponding phenomena in neuroscience and psychology / Brain area responsible}\\
 \hline
 \endfirsthead

 \hline
 \multicolumn{2}{|c|}{Continuation of Table \ref{tab:summary-table}}\\
 \hline
\textbf{Reinforcement learning concept} &
  \textbf{Corresponding phenomena in neuroscience and psychology / Brain area responsible}\\
 \hline
 \endhead

 \hline
 \endfoot

 \hline
 \multicolumn{2}{| c |}{End of Table}\\
 \hline\hline
 \endlastfoot

state value function                                            & reward expectancy in basal ganglia \cite{hikosaka_basal_2006}, prefrontal cortex \cite{wallis_heterogeneous_2010} and other areas \cite{schultz_neuronal_2000}             \\ \hline
action value function                                           & chosen value during decision making \cite{samejima2005representation, padoa-schioppa_neurons_2006, cai_heterogeneous_2011, lau2008value}               \\ \hline
multi-task learning \cite{lazaric2010bayesian}                                            & distributed reward signals \cite{seo_lateral_2009, kim_prefrontal_2008, pastor-bernier_neural_2011, so_supplementary_2010, cai_heterogeneous_2011, murdoch2018place, padoa-schioppa_neurons_2006}                                                        \\ \hline
actor-critic baseline \cite{sutton1999policy}                                          & relative values \cite{seo_behavioral_2009, kim_prefrontal_2008, pastor-bernier_neural_2011, seo_lateral_2009, cai_heterogeneous_2011}                                                                    \\ \hline
reward prediction error / TD error &
  RPE signals in dopaminergic neurons (VTA) \cite{schultz_behavioral_2006}, orbitofrontal cortex, lateral habenula, cingulate cortex, etc. \cite{matsumoto_lateral_2007, hong_globus_2008, matsumoto_medial_2007, seo_temporal_2007, sul_distinct_2010, kim_role_2009, oyama_reward_2010} \\ \hline
credit assignment problem                                       & orbitofrontal cortex \cite{iversen_perseverative_1970, schoenbaum_orbitofrontal_2002, fellows_ventromedial_2003, murray_what_2007}                                                              \\ \hline
eligibility traces \cite{sutton_reinforcement_1998}                                             & prefrontal cortex, striatum, frontal cortex \cite{barraclough_prefrontal_2004, seo_behavioral_2009, seo_lateral_2009, kim_encoding_2007, kim_role_2009, sul_distinct_2010, sul_role_2011, curtis_beyond_2010}                                       \\ \hline
experience replay \cite{mnih_playing_2013, schaul_prioritized_2016}                                              & hippocampal place cells \cite{moser_place_2008, diba_forward_2007, foster_reverse_2006, louie_temporally_2001, skaggs_replay_1996}, entorhinal cortices \cite{olafsdottir_coordinated_2016}, visual cortices \cite{ji_coordinated_2007}                      \\ \hline
episodic memory \cite{lin_episodic_2018}                                                 & instance-based models of memory \cite{gershman_reinforcement_2017, lengyel_hippocampal_nodate, bornstein_reinstated_2017, bornstein_reminders_2017}                                                   \\ \hline
temporal difference (TD) learning \cite{samuel_studies_1959, sutton_learning_1988}                              & reward prediction error hypothesis of dopamine neuron activity \cite{schultz_predictive_1998}                    \\ \hline
TD($\lambda$) \cite{sutton_reinforcement_1998}                                                  & TD model of classical conditioning \cite{montague_framework_1996}                                                \\ \hline
model-based RL                                                 & Cognitive maps \cite{PMID:18870876}; role of PFC \cite{glascher_states_2010}, hippocampus \cite{sirota_entrainment_2008, benchenane_coherent_2010, hyman_what_2011, womelsdorf_selective_2010}                                                \\ \hline
successor representations \cite{Dayan1993ImprovingGF} & neural substrates \cite{Gershman2018TheSR}, SR as between between model-free and model-based systems \cite{Akam2021WhatID, Russek2017PredictiveRC} \\ \hline
distributional TD learning \cite{bellemare_distributional_2017}                                  & distributional coding in non-RL domains \cite{pouget_probabilistic_2013, lammel_reward_2014, dabney_distributional_2018}, value coding in VTA of mice \cite{dabney_distributional_2020}                                                        \\ \hline
meta reinforcement learning \cite{schaul_metalearning_2010, wang_learning_2016, duan_rl2_2016}                                    & learning to learn \cite{harlow_formation_1949}, fast and slow learning \cite{botvinick_reinforcement_2019}, prefrontal cortex \cite{wang_prefrontal_2018, tsutsui_dynamic_2016, rushworth_choice_2008, padoa-schioppa_neurons_2006, kim_neural_1999, barraclough_prefrontal_2004}                      \\ \hline
episodic meta RL \cite{ritter_episodic_2018}                                               & cerebral cortex \cite{ritter_been_2018, santoro_meta-learning_2016, wayne_unsupervised_2018}, interaction between meta model-based control and episodic memory in human learning \cite{vikbladh_episodic_2017} \\ \hline
causality \cite{10.5555/1642718} & children perform interventions \cite{McCormack2016ChildrensUO, Sobel2010TheIO}; counterfactuals \cite{Palminteri2017ConfirmationBI} in mPFC \cite{Pischedda2020TheEO}, frontal cortex \cite{Boorman2011CounterfactualCA}, dopamine fluctuations \cite{Kishida2015SubsecondDF}; causal induction \cite{Gershman2010LearningLS, gershman_exploring_2012, soto_explaining_2014} \\ \hline
hierarchy \cite{barto_recent_2003, dietterich_hierarchical_2000, parr_reinforcement_1998, sutton_between_1999} &
  goal-directed behavior \cite{lashley_problem_1951, miller_plans_2017, newell_report_1959, anderson_integrated_2004, botvinick_doing_2004, schneider_hierarchical_2006, zacks_event_2007}, prefrontal cortex \cite{badre_cognitive_2008, botvinick_hierarchical_2008, courtney_hierarchical_2012, fuster_theory_1989, koechlin_architecture_2003, ribas2011neural, ribas2019subgoal, balaguer2016neural}, higher level representation of task context \cite{lashley_problem_1951}, mental hierarchical organization \cite{newtson_foundations_1976, zacks_event_2001}, gradual integration of skills \cite{bruner_organization_1973, fischer_theory_1980, greenfield_development_1972}, production-system based theories of cognition (ACT-R, Soar) \cite{anderson_integrated_2004} \\ \hline
options \cite{sutton_between_1999}                                                        & task representation \cite{cohen_control_1990, cooper_contention_2000, monsell_task_2003}                                                               \\ \hline
incompatibility between learning problem and temporally abstract actions \cite{botvinick_hierarchically_2009} & negative transfer problem \cite{luchins_mechanization_1942}                                                         \\ \hline
option discovery                                                & integration of causal representations \cite{gopnik_mechanisms_2004, gopnik_theory_2004, sommerville_infants_2005, sommerville_pulling_2005}, intrinsic rewards \cite{bunzeck_absolute_2006, redgrave_short-latency_2006, schultz_responses_1993}, Bayesian models \cite{tomov2020discovery, solway2014optimal, schapiro2013neural}                              \\ \hline
graph partitioning to identify bottleneck states \cite{mannor_dynamic_2004, menache_q-cut_2002, simsek_identifying_2005}&
  children integrate causal representations into a causal model \cite{gopnik_mechanisms_2004, gopnik_theory_2004, sommerville_infants_2005, sommerville_pulling_2005} \\ \hline
intrinsic motivation \cite{barto_intrinsically_2004, singh_intrinsically_2005}                                           & intrinsic rewards in dopamine driven learning \cite{bunzeck_absolute_2006, redgrave_short-latency_2006, schultz_responses_1993}                                     \\ \hline
 \end{longtable}

\section{Discussion}
\label{sec:discussion}

Reinforcement learning's emergence as a state-of-the-art machine learning framework and concurrently, its promising ability to model several aspects of biological learning and decision making, have enabled research at the intersection of reinforcement learning, neuroscience and psychology. Through this review, we have attempted to comprehensively illustrate the various classes of RL methods and validation of these methods in brain science literature. Table \ref{tab:summary-table} summarizes all the findings discussed in the paper and provides a mapping between RL concepts and evidence corresponding to them in neuroscience and psychology. \\

As detailed in earlier sections, findings in brain science have played an important role in inspiring new reinforcement learning ideas such as value functions, eligibility traces, TD learning, meta RL and hierarchical RL. Given this success, we now discuss some ways in which existing work in brain science could potentially further influence reinforcement learning research. 
\begin{itemize}
    \item \textbf{Distinct rewards:} In the brain, distinct rewards condition behavior for a single decision, and each reward might be encoded by different brain regions \cite{seo_lateral_2009, kim_prefrontal_2008, pastor-bernier_neural_2011, cai_heterogeneous_2011, so_supplementary_2010, murdoch2018place, padoa-schioppa_neurons_2006, tremblay1999relative, wallis2003neuronal}. Rewards in modern deep RL algorithms always update all parameters in the network, which is inefficient and could potentially impact learning in cases where objectives desired by two or more reward signals conflict.
    \item \textbf{Action eligibility traces:} Neurons in the orbitofrontal cortex are believed to encode relationships between actions and their corresponding outcomes \cite{barraclough_prefrontal_2004, seo_lateral_2009, kim_role_2009, sul_distinct_2010}. RL algorithms predominantly use only state eligibility traces, which is sufficient when the same actions are available at every state. However, in real-world cases where different states allow different sets of actions, a trace for actions might be useful.
    \item \textbf{`Imagination' replay:} Experience replay in deep RL models tend to replay past-experienced sequences of states in the order in which they occurred \cite{mnih_playing_2013, schaul_prioritized_2016}. Recent observations in neuroscience suggest that hippocampal replay might replay states based on the expected sequence according to an internal world model \cite{liu_human_2019}. This event-based replay is currently unexplored in RL literature.
    \item \textbf{Action vigor:} Work on the role of motivation in animal behavior observes a strong correlation between motivation and action vigor \cite{skinner1988operant}. Though modern RL algorithms do model motivation using intrinsic motivation and curiosity \cite{barto_intrinsically_2004, singh_intrinsically_2005}, most RL environments do not allow agents to adjust the rate at which actions are applied. This currently prevents potentially interesting analysis on vigor, caution and motivation in RL agent behavior.
    \item \textbf{Grounded language learning}: Language plays an important role in many aspects of human learning such as exploration and forming internal representations. Recent work in computational linguistics also emphasizes the role of pragmatic communication in human representation learning \cite{cohn2018incremental}. Recent work in RL has embraced language and has made strides towards proposing algorithms and challenges that present language as an important tool for learning \cite{colas2020language, narasimhan2018grounding}.
    \item \textbf{Social learning}: A key aspect of human learning is our ability to learn via social interaction. Theory of Mind (TOM) \cite{premack1978does} involves the ability to understand others by attributing mental beliefs, intents, desires, emotions, and knowledge to them. Recent studies show that neural signals carry social information, as reviewed comprehensively by Insel and Fernald \cite{insel2004brain}. Social situations are much more complex which may not be completely expressible from a single reward value. 
    \item \textbf{Modularity}: The human brain is known to be very modular in nature. Specific regions specialize in specific roles such as the occipital lobe deals primarily with vision signals, the temporal lobe deals primarily with auditory signals, etc. Modularity makes learning systems flexible, adaptable and allow quick transfer of decision making knowledge allowing humans to be versatile at learning. Although work exists on modular reinforcement Learning \cite{simpkins2019composable, sprague2003multiple}, most current RL algorithms and computational models lack modularity and are unable to adapt to new tasks, domains, etc quickly, as there is a lack of formalism guiding the definition and use of modularity. It is a core component for both our internal world models as well as in sequential decision making \cite{fodor1983modularity}. In the former case, modularity leads to building parts that may specialize in a domain, for instance, processing a particular type of sensory signal. In the latter case, modularity results in efficient adaptation to new tasks, domains, etc. Recently, Cheng et. al. \cite{chang2021modularity} built principles for modular decision making in the model-free setting by establishing principles to identify credit assignment algorithms which allow independent modification of components. 
\end{itemize}

Leveraging and implementing ideas from neuroscience and psychology for RL could also potentially help address the important issues of sample efficiency and exploration in modern-day algorithms. Humans can learn new tasks with very little data (Figure \ref{fig:humansvsrl}). Moreover, they can perform variations of a learned task (with different goals, handicaps etc.) without having to re-learn from scratch \cite{lake_ullman_tenenbaum_gershman_2017, tsividis_human_nodate}. Unlike deep RL models, humans form rich, generalizable representations which are transferable across tasks. Principles such as compositionality, causality, intuitive physics and intuitive psychology have been observed in human learning behavior, which if replicated or modeled in RL, could produce large gains in sample efficiency and robustness. \cite{lake_ullman_tenenbaum_gershman_2017} \\

Studies in neuroscience and psychology have also been motivated by ideas originally developed in RL literature. For instance, the temporal difference model made several important predictions about classical conditioning in animals \cite{sutton1981toward, sutton1990time}, some of which were verified experimentally only later \cite{kehoe1987temporal}. Studies into the possibility of distributional coding in the brain \cite{dabney_distributional_2020} were motivated by the distributional TD model developed a few years before \cite{bellemare_distributional_2017}. Another example is meta reinforcement learning which according to Wang et al. \cite{wang_prefrontal_2018} was first observed as an emergent phenomenon in recurrent networks trained using an RL algorithm, which inspired research into the possibility of the prefrontal cortex encoding similar behavior. All of these examples suggest that reinforcement learning is a promising model for learning in the brain and therefore that experimenting with RL models could yield predictions that motivate future research in neuroscience and psychology.

\section{Acknowledgements}
We would like to thank the Cognitive Neuroscience Lab, BITS Pilani Goa for supporting and enabling this review.

\bibliography{mybibfile}

\end{document}